\documentclass[10pt,twocolumn,letterpaper]{article}

\usepackage{iccv}              

\usepackage{amsmath}
\usepackage{bbding}
\usepackage{multirow}
\usepackage{booktabs}
\usepackage{makecell}

\definecolor{iccvblue}{rgb}{0.21,0.49,0.74}
\usepackage[pagebackref,breaklinks,colorlinks,allcolors=iccvblue]{hyperref}

\title{Pathology Image Restoration via Mixture of Prompts}

\author{
    Jiangdong Cai$^{1*}$ \quad Yan Chen$^{1*}$ \quad Zhenrong Shen$^{2}$ \quad Haotian Jiang$^{1}$ \quad Honglin Xiong$^{1}$ \\
    Kai Xuan$^{4}$ \quad Lichi Zhang$^{2}$ \quad Qian Wang$^{1,3 \dag}$ \\ 
    \textsuperscript{1}School of Biomedical Engineering \& State Key Laboratory of\\Advanced Medical Materials and Devices, ShanghaiTech University\\
    \textsuperscript{2}School of Biomedical Engineering, Shanghai Jiao Tong University\\
    \textsuperscript{3}Shanghai Clinical Research and Trial Center\\
    \textsuperscript{4}Nanjing University of Information Science and Technology \\
}

\begin{document}

\maketitle

\begin{abstract}

In digital pathology, acquiring all-in-focus images is essential to high-quality imaging and high-efficient clinical workflow. 
Traditional scanners achieve this by scanning at multiple focal planes of varying depths and then merging them, which is relatively slow and often struggles with complex tissue defocus. 
Recent prevailing image restoration technique provides a means to restore high-quality pathology images from scans of single focal planes. 
However, existing image restoration methods are inadequate, due to intricate defocus patterns in pathology images and their domain-specific semantic complexities.
In this work, we devise a two-stage restoration solution cascading a transformer and a diffusion model, to benefit from their powers in preserving image fidelity and perceptual quality, respectively.
We particularly propose a novel \textbf{mixture of prompts} for the two-stage solution.
Given initial prompt that models defocus in microscopic imaging, we design two prompts that describe the high-level image semantics from pathology foundation model and the fine-grained tissue structures via edge extraction. 
We demonstrate that, by feeding the prompt mixture to our method, we can restore high-quality pathology images from single-focal-plane scans, implying high potentials of the mixture of prompts to clinical usage. Code will be publicly available at \url{https://github.com/caijd2000/MoP}.

\end{abstract}    
\section{Introduction}
\label{sec:intro}

For accurate and high-throughput disease diagnosis using pathology images, a fundamental prerequisite is to capture high-quality and all-in-focus images \cite{wright2020effect,bian2020autofocusing}.
However, scanners face challenges balancing quality and speed due to out-of-focus blur. 
On one hand, the thickness of bio-samples inherently complicates achieving an all-in-focus view, impeding pathologists and AI-based diagnostics.
Though technologies of auto-focusing or assisted-focusing have helped ease the defocus, image degradation remains on uneven and non-coplanar tissues.
On the other hand, the mainstream solution of multi-focus image fusion (MFIF)  \cite{liu2020multi,bhat2021multi} involves capturing images at various focal planes and fusing them into an all-in-focus composite, imposing a considerable time burden.
Hence, there is a pressing need for a feasible pathology image acquisition technique, i.e., to restore high-quality images from a single scans.

\begin{figure}
    \centering
    \includegraphics[width=\linewidth]{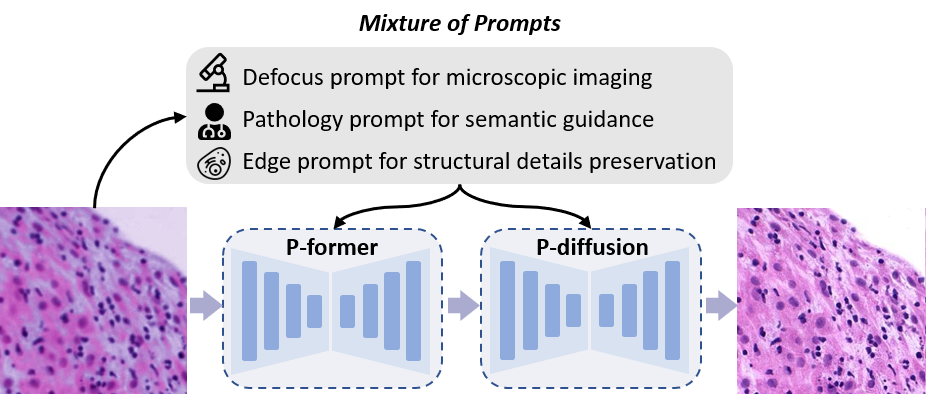}
    \caption{Our method addresses key issues in pathology image restoration by extracting a mixture of prompts. These prompts are utilized to guide a two-stage restoration process, consisting of a coarse-stage P-former and a fine-stage P-diffusion.}
    \label{fig:intro}
\end{figure}

Though recent image restoration methods have achieved great success in real-world image restoration \cite{zamir2022restormer,wang2022uformer,li2023efficient,quan2023single,luo2023image,luo2023controlling,lin2023diffbir}, they struggle to ensure both pixel-level fidelity and perception-level quality for pathology images. 
The substantial discrepancy between microscopy datasets and real-world datasets has been identified in the insightful analysis conducted by Zhang et al. (2023) \cite{zhang2024unified}. 
Unlike natural images with larger pixel chunks and fewer instances, pathology images contain numerous densely distributed and morphologically diverse cell nuclei and a multitude of high-frequency textures, folds, and wrinkles. 
These intricate details are almost entirely lost under severe defocus, which presents a formidable challenge to current methodologies.
\begin{itemize}[leftmargin=2em]
\item \textbf{Network-based methods}: Network-based methods without diffusion are optimized by minimizing the per-pixel differences, often resulting in over-smoothing images. 
Due to the relatively uniform color distribution in pathological images compared to real-world data, models tend to achieve lower pixel-wise errors by fitting smooth mappings, rather than prioritizing the reconstruction of high-frequency textures, which are more perceptually significant.
While perceptual losses \cite{ledig2017photo} can help reduce blurry outputs, they may also introduce unwanted artifacts.
\item \textbf{Diffusion-based methods}: 
Early Diffusion-based methods \cite{lin2023diffbir,xia2023diffir} alone struggle to effectively retain critical details, such as cell nuclei and edges, due to the intricate and densely packed nature of pathological instances.
Recently, the residual diffusion paradigm \cite{luo2023image,luo2023controlling,liu2024residual,yue2024efficient,zhenning2023resfusion} has shown promise in detail preservation for image restoration tasks. 
However, these models still face significant challenges when dealing with severely blurred inputs, often failing to ensure precise recovery of details.
This issue arises because diffusion models, while improving perceptual performance, may not accurately estimate noise in the presence of severe degradations, ultimately affecting the fidelity of the restored image.

\end{itemize}

To overcome the limitations of each paradigm, a straightforward approach is to combine network-based and diffusion-based methods to leverage the strengths of both.
In our initial experiments, this hybrid method has managed to strike a balance between perceptual clarity and pixel-wise fidelity. 
However, their simple combination cannot result in a significant performance breakthrough because of the failure to address the underlying \textbf{insufficient prior}.
Restoring cellular textures in defocused images necessitates a deep understanding of the pathology domain.
Unfortunately, existing pathology image restoration datasets are limited in size and only provide pixel-level annotations, lacking higher-level semantic information.
Consequently, it remains challenging to incorporate comprehensive prior knowledge in pathology effectively \cite{bulten2022artificial,kang2023benchmarking}.
This limitation restricts the image restoration network’s ability to go beyond simple local pixel mapping, hindering its capability to truly understand and restore the image content.

Building upon the identified challenges, we propose the \textbf{mixture of prompts}, leveraging pre-extracted prompts to address key issues effectively. Moreover, we design the two-stage restoration methods, combining network-based method P-former (Prompt-driven Restormer) and diffusion-based method P-diffusion (Prompt-driven diffusion) to optimize the utilization of these prompts.
Our methods, which surpass the capabilities of current methods with significant improvements, can be summarized as follows.

\begin{itemize}[leftmargin=2em]

\item \textbf{Defocus-aware Prompts Generation}: 
Our methodology initiates from three distinct prompts: the pathology prompt, sourced from a pathology foundation model; the defocus prompt, derived from a pre-trained defocus encoder; and the edge prompt, obtained via Canny edge detection.
Since degraded images alone cannot generate reliable pathology and edge prompts, the defocus prompt enhances the pathology prompt and estimates confidence for weighting the edge prompt regionally.
These pre-extracted prompts contain both pathology knowledge and reliable structure, providing sufficient priors for image restoration.
\item \textbf{P-former}: In the coarse stage, we introduce P-former, which uses the pathology prompt for semantic guidance.
Leveraging the gating mechanisms of the Mixture of Experts (MoE)
\cite{jacobs1991adaptive,shazeer2017outrageously}, P-former adeptly handles image restoration across diverse tissue types and organs, enabling a more focused and precise restoration process.
\item \textbf{P-diffusion}:  In the fine stage, we present P-diffusion to refine images and optimize high-frequency details. 
This stage integrates the restored pathology prompt via cross-attention in the lower layers of the conditional U-Net, and incorporates the weighted edge prompt into the input layers, ensuring the restoration retains critical details. 
This two-stage approach systematically addresses the complexity of image restoration, with each stage focusing on specific aspects to achieve comprehensive restoration.
\end{itemize}

\section{Related Work}
\label{sec:formatting}

\subsection{Image Restoration}
Image restoration involves the process of enhancing a low-quality image to restore its high-quality version.
Researchers focus on the design of networks for better feature fusion and computational efficiency in the early years.
For example, the talent of Transformer has been leveraged by many \textit{network-based methods} including Restormer \cite{zamir2022restormer}, Uformer \cite{wang2022uformer}, and GRL \cite{li2023efficient}. 
In the realm of out-of-focus blurring, INIKNet \cite{quan2023single} aims to enhance defocus kernel representations guided by the point spread function.
However, the conventional training paradigm typically results in over-smoothing images.

While diffusion models have gained prominence in image generation, they have also catalyzed remarkable advancements of \textit{diffusion-based restoration}. 
For instance, DiffBIR \cite{lin2023diffbir} adopts a two-step approach: pre-training a restoration module to handle various degradations and then leveraging latent diffusion models for realistic image restoration.
DiffIR (Xia et al. \cite{xia2023diffir}) employs an efficient diffusion model that includes a compact IR prior extraction network and dynamic IR transformer.
However, early diffusion-based methods were capable of producing visually appealing outputs but could not ensure consistency in details with the target images. 
This limitation has been eased with the advent of residual diffusion models \cite{liu2024residual}.
IR-SDE \cite{luo2023image} is a mean-reverting Stochastic Differential Equation that transforms a high-quality image into a degraded counterpart as a mean state with fixed Gaussian noise.
DA-CLIP \cite{luo2023controlling} integrates pre-trained vision-language models based on IR-SDE.
ResShift \cite{yue2024efficient} establishes a Markov chain that facilitates the transitions between the high-quality and low-quality images by shifting their residuals, and Resfusion \cite{zhenning2023resfusion} determines the optimal acceleration step and maintains the integrity of existing noise schedules.
Nevertheless, due to the lack of prior knowledge, the residual diffusion model still falls short in addressing scenes that are excessively blurred.

Compared to natural images, pathological image restoration is indeed a relatively niche field. 
While we identified several related studies \cite{zhang2024unified,rong2023enhanced,ma2022unsupervised}, most of them did not provide open-source code or simply applied techniques from the natural image domain.
Broadening our focus to medical imaging, we found RAT \cite{yang2024rat} to be a noteworthy work, which leverages SAM \cite{kirillov2023segment} to extract medical-specific priors, achieving state-of-the-art (SOTA) results in pathological image super-resolution.

\subsection{Extraction of Priors for Downstream Tasks } 

Foundation models designed for robust feature extraction and adaptability have revolutionized several industries. 
These models typically benefit from large-scale datasets~\cite{deng2009imagenet, lin2014microsoft}, and advanced pre-training techniques like masked image modeling (MIM)~\cite{he2022masked,zhou2022mimco} and contrastive learning~\cite{he2020momentum}.
Initially, fine-tuning pre-trained models was common practice to enhance performance in downstream tasks such as classification, detection, and segmentation \cite{marcelino2018transfer,zhuang2020comprehensive,zhao2023clip}. 
In pathology, foundation models~\cite{nechaev2024hibou,lu2024visual,hoptimus0,huang2023visual} have made significant strides, enabling more accurate, efficient, and consistent diagnoses. 
For instance, UNI~\cite{chen2024towards} introduced capabilities like resolution-agnostic tissue classification, few-shot slide classification, and generalization in cancer subtyping classification.
Prov-GigaPath~\cite{xu2024whole} was trained on 1.3 billion pathology image tiles using the innovative GigaPath architecture, achieving state-of-the-art performance on 25 out of 26 tasks by modeling ultra-large contexts.
Despite these advancements, pathology-based models have yet to be utilized for optimizing pathology image restoration fully. 

Recently, inspired by advancements like stable diffusion~\cite{rombach2022high}, features extracted from pre-trained models have increasingly been utilized as prompts for downstream applications. 
In the context of image restoration, pre-trained models are often employed to introduce representations from various modalities as guidance \cite{luo2023controlling,xu2024boosting,liu2024diff,qi2023tip,lin2024improving,chen2023image}.
Other research efforts have concentrated on enhancing these representations with degradation-specific knowledge through prompt learning, aiming to achieve better generalization in image restoration tasks \cite{potlapalli2024promptir,li2024promptcir,guo2024onerestore}. 
These prompt-based approaches have primarily been used to differentiate between various types of degradation for all-in-one restoration. 
However, while these methods provide valuable insights, they are not directly equipped to handle the complexities posed by the intricate environments found in pathology images.

\begin{figure}[ht]
    \centering
    \includegraphics[width=1\linewidth]{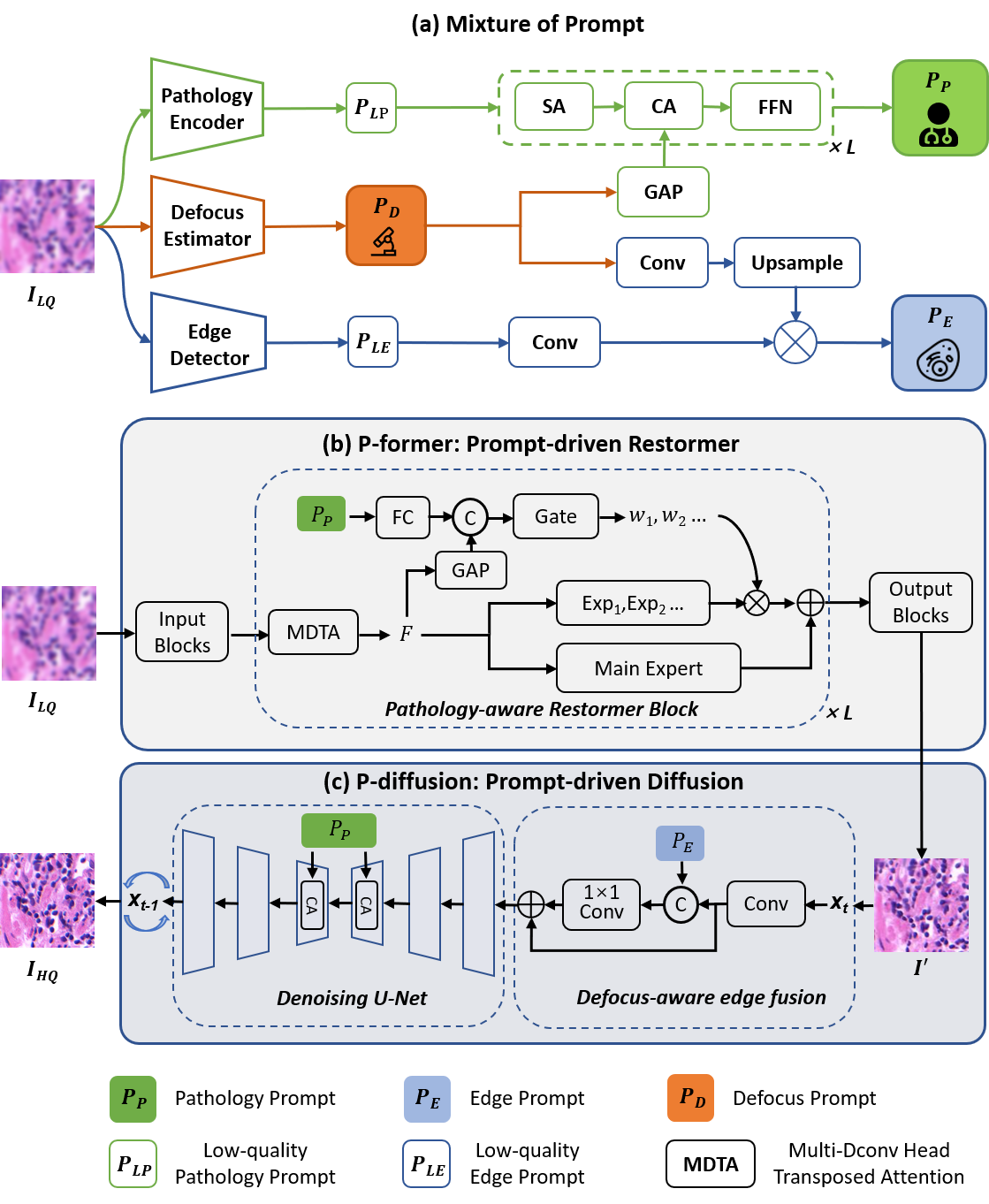}
    \caption{The framework of our method. (a) Mixture of prompts. The defocus prompt $P_D$ is utilized to restore the pathology prompt $P_P$ and weight the edge prompt $P_E$ from the low-quality image. (b) P-former uses the pathology prompt $P_{P}$ to enhance Restormer \cite{zamir2022restormer} blocks within MoE. (c) P-diffusion is constrained structurally and semantically by the edge prompt $P_{E}$ and  $P_{P}$.}
    \label{fig:pipeline}
\end{figure}

\section{Method}
\label{sec:method}


As illustrated in Figure \ref{fig:pipeline}, our overall process involves the extraction and utilization of prompts using \textbf{mixture of prompts} approach. This guides both the coarse-stage \textbf{P-former} (from the low-quality image $I_{LQ}$ to the coarsely restored image $I^{'}$) and the fine-stage \textbf{P-diffusion} (from $I^{'}$ to the high-quality image $I_{HQ}$).

To effectively address the complexities of pathology image restoration, we generate prompts from the low-quality image $I_{LQ}$, by leveraging prior knowledge as detailed in Figure \ref{fig:pipeline}(a). 
Our prompt design incorporates essential information about pathological microscopic images, focusing on defocus awareness, pathology understanding, and structural preservation.
First, a pre-trained defocus encoder creates the defocus prompt $P_D$, the pathology foundation model extracts the low-quality semantics prompt $P_{LP}$ (due to low-quality input image), and an edge detector produces the low-quality edge prompt $P_{LE}$.
$P_D$ is then used to refine both $P_{LP}$ and $P_{LE}$, resulting in the high-quality pathology prompt $P_P$  and the edge prompt $P_E$.

As shown in Figure \ref{fig:pipeline}(b-c),  our two-stage solution leverages conventional restoration models for structural recovery and diffusion models for rendering realistic fine details.
During the coarse restoration process, prompts guide the workflow in distinct ways: $P_{P}$ governs MoE within P-former, enabling the decoupling of restoration tasks across various semantic scenarios and their processing by specialized experts. 
Furthermore, the fine P-diffusion enhances images both semantically through cross-attention \cite{rombach2022high} and structurally by integrating the defocus-aware edge fusion with $P_{E}$.

The following sections of our paper provide an in-depth explanation of our proposed framework and its components.

\subsection{Mixture of Prompts}


Despite the wealth of information available through edge detectors and pre-trained pathology foundation models, the defocus blur inherent in low-quality input images often compromises the accuracy of this information.
To address this challenge, we propose extracting a defocus prompt using a pre-trained defocus encoder. This defocus prompt is then combined with the low-quality pathology prompt and edge prompt to produce high-quality prompts, as detailed in Figure \ref{fig:pipeline}(a).

\textbf{Defocus Prompt ($P_D$).} 
In the context of pathological imaging, it is common practice to scan the same field of view at pre-set depths for multiple times. 
The degree of defocus blur is directly related to the distance to the focal plane that captures the sharpest image. 
Consequently, the task of defocus estimation can be effectively approached as a quantitative prediction of this distance. 
Furthermore, the contrast transfer function (CTF) provides a mathematical value of how optical aberrations affect the sample's image \cite{wade1992brief,martin1995experimental,reimer2013transmission,kirkland1998advanced}. 
The CTF value is not only instrumental in identifying the optimal focus layer but also aids significantly in the training of our defocus estimation model.

Specifically, Defocus Estimator $D_{\phi }$ in Figure \ref{fig:pipeline}(a) is pre-trained to predict the focal distance and CTF value meanwhile in a single shot, the extracted features of which can represent the defocus level of $I_{LQ}$. 
Defocus Estimator takes ResNet-34 \cite{he2016deep} as the backbone, and RandStainNA \cite{shen2022randstainna} is leveraged for stain normalization $S( \cdot)$ to diversify stain styles in the input image.
The predicted distance and CTF value are produced by: $\{\widehat{d}, \widehat{c}\} = D_{\phi}(S(X))$, and
the training of the Defocus Estimator follows:

\begin{equation}
    \begin{aligned}
     L_{def} =  \Vert d-\widehat{d} \Vert _1 + \Vert c-\widehat{c} \Vert _1,
    \label{con:defocus}
    \end{aligned}
\end{equation}
where $d$ and $\hat{d}$ are the ground truth and prediction of defocus distance, $c$ and $\hat{c}$ are the ground truth and prediction of CTF value. 
The features before pooling are ultimately defined as the defocus prompt $P_D$, helping restore the pathology prompt and reweight the edge prompt.

\textbf{Defocus-aware Pathology Prompt ($P_P$).} Pathology images display a wide array of intricate textures and structures across different organs and tissues.
Recovering these cellular textures requires substantial domain-specific knowledge, which can be difficult for restoration models to capture due to limitations in data and annotations.
Consequently, pathology foundation models~\cite{nechaev2024hibou,lu2024visual,hoptimus0,huang2023visual} are employed for generating pathology prompts.
However, these models are typically pre-trained on high-quality images, leading to potential biases when extracting information from low-quality pathological images. 
To address this issue, we introduce the use of $P_D$ to correct and refine the extracted information.

As shown in Figure \ref{fig:pipeline}(a), the pathology foundation model extracts low-quality pathology prompt, denoted as $P_{LP}$, from low-quality images.
To refine $P_{LP}$ and bring it closer to the high-quality pathology prompt $P_{HP}$ extracted from the ground truth images, we utilize the defocus prompt.
We have designed a straightforward network consisting of $L$ stacked transformer blocks to achieve this. 
Each block includes a self-attention block, a cross-attention mechanism that integrates $P_{LP}$ and $P_D$, and a Feed-Forward Network (FFN). 
This process can be described as follows:

\begin{equation}
    \begin{aligned}
     P_{P} = R_{p}(P_{LP},P_{D}), \\
     L_{P} = \Vert P_{P}-P_{HP} \Vert _1,
     \end{aligned}
    \label{con:restore path}
\end{equation}
where $R_{p}$ denotes the transformer-based network for prompt restoration. Prov-GigaPath\cite{xu2024whole} is used to extract pathology prompt in our setting. 

\textbf{Defocus-aware Edge Prompt ($P_E$).} In addition to macro-level semantic guidance, structural constraints play a crucial role in restoring pathology images.
The diffusion process is prone to introducing structural distortions, which can compromise vital diagnostic information. 
To mitigate this risk, edge detection—a commonly used technique for extracting structural information—can be employed to generate an edge prompt that guides the diffusion process.

One main challenge in applying low-quality edge prompts is the instability caused by defocus blur. 
$P_D$ encapsulates information regarding the blurriness across various image regions, indicating which edges are reliable. 
Thus, we weight the edges extracted from the image using information from the defocus prompt.
 As depicted in Figure \ref{fig:pipeline}(a), we first extract the binary edge map from the low-quality image using the Canny operator and then upscale it with a convolutional layer.
 The feature map from the defocus encoder is convolved and upsampled to match the dimensions of the edge map. 
 This processed map is then element-wise multiplied with the edge features to obtain the defocus-weighted edge prompt $P_{E}$.


\subsection{P-former}P-former is designed for the coarse restoration of pathological images.
Given the variations in thickness, density, and contrast among different organ tissues, the forms of degradation they exhibit can differ significantly. 
To overcome this, P-former harnesses the combined power of a pathology foundation model and an MoE architecture. 
The pathology prompt $P_{P}$ enables the model to preemptively align with specific pathological contexts, allowing a tailored assembly of experts to address the nuanced restoration tasks. 
This integrated approach significantly enhances the capacity of P-former to effectively manage the broad spectrum of degradation patterns encountered in pathological imaging.

The architecture of the P-former is illustrated in Figure \ref{fig:pipeline}(b). 
P-former is based on the widely adopted Restormer \cite{zamir2022restormer}, with the significant enhancement of incorporating MoE into the Gated-Dconv Feed-Forward Network (GDFN). 
The core innovation in P-former is the use of distinct depth-wise convolution (DConv) layers as separate experts, with a router that adaptively adjusts the weights assigned to these experts.
In the gating section, the pathology prompt $P_{P}$ is concatenated with the pooled image features $F$ and fed into the router.
The router then predicts the weights $w_{1...n}$ via a linear layer, normalizing them with the softmax function.
In the expert section, we create a main expert with a constant weight of 1 and $n$ sub-experts, which are weighted by $w_{1...n}$.  
The weighted outputs are summed and passed through a GeLU activation function before being combined with the output of the main expert.
The process can be formulated as:
\begin{equation}
\begin{aligned}
w_i &= \text{softmax}(\text{Linear}(\text{concat}( GAP(F), P_{P})))_i, \\
F_o &= E_0(F) + \text{GeLU}(\sum_{i=1}^{n} w_i \cdot E_i(F)),
\end{aligned}
\end{equation}
where $F$ and $F_o$ are the input and output feature, $GAP(\cdot)$ is global average pooling, $E_{0}$ is the main expert and $E_1, ..., E_n$  are the sub-experts. The number of experts $n$ is set to 3 for all blocks.

\subsection{P-diffusion} 
Diffusion-based methods often tend to produce unrealistic artifacts. 
To address this issue, we design P-diffusion, which leverages a pathology prompt to provide semantic guidance and an edge prompt to impose structural constraints during the generation process.
It is important to note that the edge prompt is only utilized in P-diffusion, rather than P-former, because structural information is difficult to preserve in noisy states but can be easily extracted using network-based methods.

P-diffusion is based on ResShift \cite{yue2024efficient}, which establishes a Markov chain that enables efficient transitions between high-quality and low-quality images by manipulating their residuals. 
The transition distribution of ResShift can be formulated as:
\begin{equation}
    \begin{aligned}
    q\left(\boldsymbol{x}_{t} \mid \boldsymbol{x}_{t-1}, \boldsymbol{I}_{LQ}\right)=\mathcal{N}\left(\boldsymbol{x}_{t} ; \boldsymbol{x}_{t-1}+\alpha_{t} \boldsymbol{e}_{0}, \kappa^{2} \alpha_{t} \boldsymbol{I}\right),
    \end{aligned}
\end{equation}
where $x$ is the state of the diffusion process, the residual is denoted by $e_0=I_{LQ}-I_{HQ}$, $\alpha _ {t}$ is a time-dependent parameter, and $\kappa$ is a hyperparameter controlling the noise variance.
We enhance the estimation of the posterior distribution by introducing semantic and structural conditions $P_{P}$ and $P_{E}$.
The objective function follows:

\begin{equation}
    \begin{aligned}
    \min_{\theta} \sum_{t} w_{t} \| f_{\theta}(x_{t}, I_{LQ}, t , P_{P} , P_{E} ) - I_{HQ} \|_{2}^{2},
    \end{aligned}
\end{equation}
where $f_{\theta}$ is a U-Net \cite{ronneberger2015u} like architecture with parameter $\theta$, aiming to predict $I_{HQ}$.

As detailed in Figure \ref{fig:pipeline}(c), $P_{P}$ is integrated into the diffusion process via cross-attention mechanism \cite{rombach2022high}, a widely adopted technique for leveraging semantic guidance. 
To manage the computational cost associated with applying attention to high-resolution features, cross-attention blocks are selectively employed to enhance the deep features in the bottom U-Net blocks.
The Defocus-aware Edge Fusion integrates the edge prompt $P_{E}$ at the input stage of the U-Net.
The noisy state $x_t$ is subjected to $3\times3$ convolution, producing the shallow features, and then concat with $P_{E}$ and fused by $1\times1$ convolution.

\section{Experiments}
\label{sec:experiment}

\subsection{Datasets and Implementation Details}

\textbf{Datasets.} 
Collaborating with a clinical institute, we collected an in-house dataset specifically to validate the effectiveness of our method on histopathological images.
For the restoration model training, we utilized 100 histopathology slides from 100 distinct patients with ethical approval. 
The datasets are diverse to ensure better generalization, covering multiple tissue types (e.g., STR, DEB, CRC) from multiple organs (e.g., uterus, skin, colon).
Each slide was scanned across 13 focal planes, with an equidistant spacing $0.8\mu m$ between adjacent planes. 
We randomly selected 10 fields of view from each slide, each with a resolution of 3072$\times$4096 pixels.
All focal planes were used as inputs, while their fused images served as the ground truths, resulting in a total of 13,000 input-output pairs.
The slides were pre-allocated into training, validation, and test sets in an 8:1:1 ratio based on the origin of their corresponding patients.
Training and inference were conducted on cropped patches of 256$\times$256 pixels, while all metrics were computed at slide level to better resemble the display format used in routine pathologists' diagnoses.
The defocus-aware restoration was trained on pairs of prompts from the corresponding low-quality and high-quality images.
To pre-train the defocus estimator, we collected an additional 50 histopathology slides.

To further validate the versatility of our method, we conducted experiments on a public cervical cytopathology dataset, 3DHistech \cite{geng2022cervical}. 
This dataset collects 108,065 defocused and focused patches of 256$\times$256 pixels from 5 slides. 
For the restoration task, we adhered to the original data partitioning. 
To enable downstream task evaluation, collaborating pathologists provided additional annotations indicating whether cervical cells were abnormal (i.e., the risk of cervical cancer). 
This annotated dataset, with 86 positive and 409 negative patches, is exclusively from the test set and will be released along with our code.

\textbf{Implementation.} 
We employed the same training strategy on both datasets.
All experiments were optimized using Adam and conducted on an Nvidia A100 GPU. 
\textit{Defocus-aware prompt restoration} is trained for 500 epochs with batch size of 16 and initial learning rate of $1\times 10^{-4}$ and, being decreased by the StepLR scheduler with $\gamma=0.98$.
Defocus estimator is trained for 100 epochs with batch size of 16 and initial learning rate of $1\times 10^{-4}$, being decreased by the StepLR scheduler with $\gamma=0.95$.
During the \textit{coarse stage} training, P-former was trained for 300 epochs with batch size of 8 and an initial learning rate of $1\times 10^{-4}$, being decreased by the StepLR scheduler with $\gamma=0.98$.
For the \textit{fine stage} training, P-Diffsion is optimized with batch size of 8 for 100,000 steps, including a warmup phase of 10,000 steps and cosine decay subsequently.
The number of diffusion steps is set to 4.


\textbf{Comparison Methods.} 
For the in-house dataset, our method is compared with three kind of methods: 
(1) Network-based methods including Restormer \cite{zamir2022restormer}, Uformer \cite{wang2022uformer}, GRL \cite{li2023efficient}, INIKNet \cite{quan2023single}, NRKNet \cite{quan2023neumann}, RAT \cite{yang2024rat};
(2) Diffusion-based methods including IR-SDE \cite{luo2023image}, DA-CLIP \cite{luo2023controlling}, Resfusion \cite{zhenning2023resfusion}, and ResShift \cite{yue2024efficient}. 
(3) Coarse-to-fine methods combining Restormer \cite{zamir2022restormer} as the coarse stage and different diffusion-based methods as the fine stage. 
For 3DHistech \cite{geng2022cervical},  we directly utilized the results reported in the paper of MPT \cite{zhang2024unified}.

\begin{table}[h]
\centering
\resizebox{1.0\columnwidth}{!}{
\begin{tabular}{c|cc|c}
\hline
\multirow{2}{*}{Method} & \multicolumn{2}{c|}{Distoration} & \multicolumn{1}{c}{Perceptual}  \\ \cline{2-4}
 & PSNR$\uparrow$ & SSIM$\uparrow$ & LPIPS$\downarrow$  \\ \hline
Uformer \cite{wang2022uformer} & 27.63 & 0.8219 & 0.1898  \\
GRL \cite{li2023efficient} & 28.46 & 0.8560 & 0.1248  \\
INIKNet \cite{quan2023single} & 27.59 & 0.8355 & 0.1196  \\
NRKNet \cite{quan2023neumann} & 28.27 & 0.8474 & 0.1590  \\
RAT \cite{yang2024rat} & 28.88 & 0.8607 & 0.1303   \\
Restormer \cite{zamir2022restormer} & 28.85 & 0.8613 & 0.1290  \\
\textbf{P-former (ours)} & \textbf{29.14} & \textbf{0.8699} & \textbf{0.1148}  \\ \hline
Resfusion \cite{zhenning2023resfusion} & \textbf{27.35} & 0.8161 & 0.1387  \\
IR-SDE \cite{luo2023image} & 13.62 & 0.6696 & 0.3046  \\
DA-CLIP \cite{luo2023controlling} & 18.78 & 0.6743 & 0.2246  \\
ResShift \cite{yue2024efficient} & 26.65 & 0.7970 & 0.0972  \\
\textbf{P-diffusion (ours)} & 27.27 & \textbf{0.8193} & \textbf{0.0852}  \\ \hline
\begin{tabular}[c]{@{}c@{}}Restormer +  Resfusion\end{tabular} & 28.27 & 0.8504 & 0.1285  \\
\begin{tabular}[c]{@{}c@{}}Restormer + IR-SDE\end{tabular} & 26.80 & 0.8052 & 0.0858  \\
\begin{tabular}[c]{@{}c@{}}Restormer + DA-CLIP\end{tabular} & 27.98 & 0.8274 & 0.0812  \\
\begin{tabular}[c]{@{}c@{}}Restormer + ResShift\end{tabular} & 27.80 & 0.8358 & 0.0809  \\
\begin{tabular}[c]{@{}c@{}}\textbf{P-former + P-diffusion (ours)}\end{tabular} & \textbf{28.91} & \textbf{0.8578} & \textbf{0.0735}  \\ \hline
\end{tabular}
}
\caption{Quantitative comparison on the in-house dataset. We categorize the \textbf{network-based}, \textbf{diffusion-based}, and \textbf{two-stage} methods into separate sections, with the best results in each category highlighted in \textbf{bold}.}
\label{tab:comparison}
\end{table}
\subsection{Comparison Results}

\begin{figure*}[h]
    \centering
    \includegraphics[width=1\linewidth]{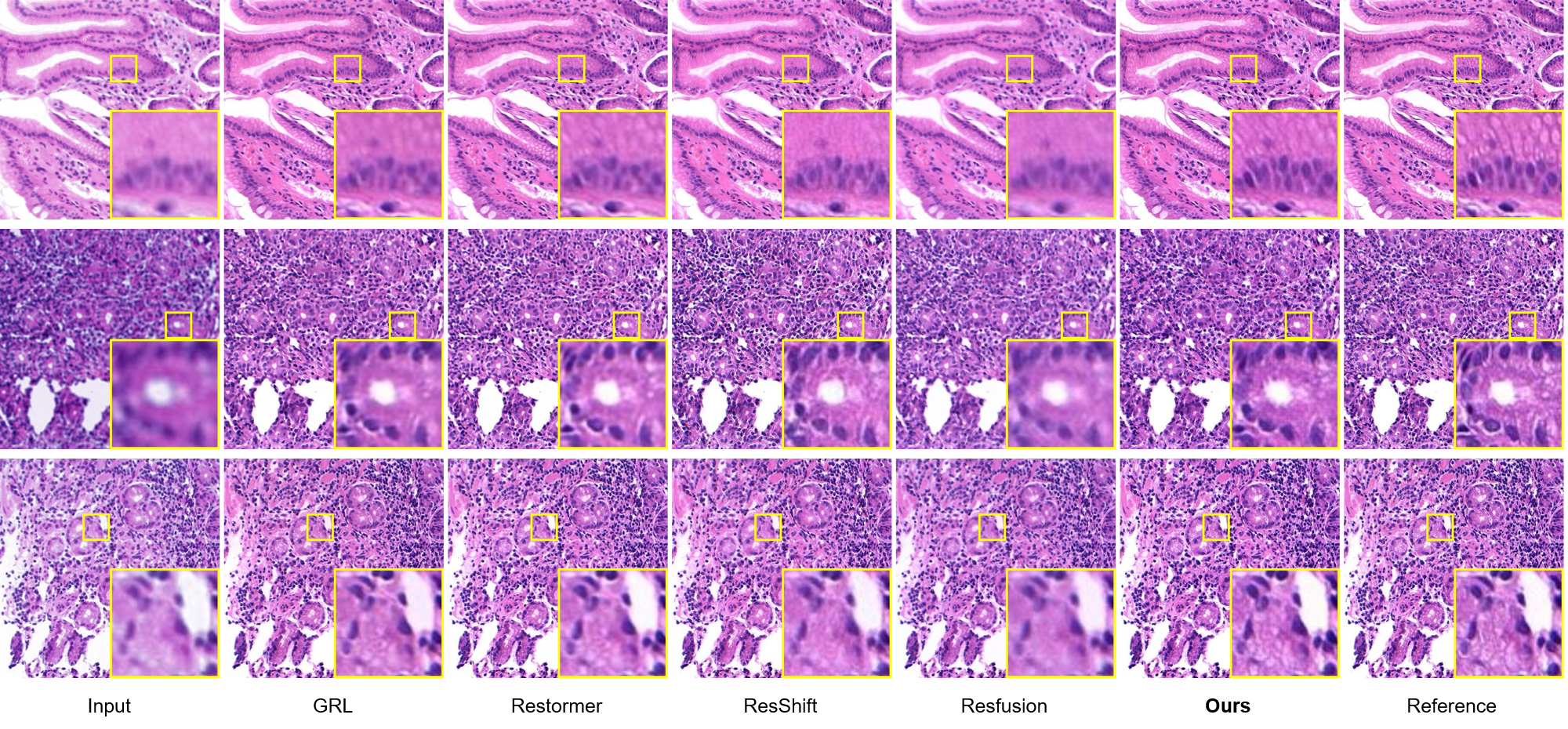}
    \caption{Visual comparison between different methods on the in-house dataset. The distinguished regions within the yellow bounding boxes are zoomed in at the bottom right.}
    \label{fig:visualization}
\end{figure*}

\textbf{Performance on In-house Histology Dataset.} 
Based on this evaluation shown in Table \ref{tab:comparison}, several significant conclusions can be drawn as follows:
\romannumeral1) Network-based methods perform well in terms of PSNR and SSIM. However, their tendency to produce overly smooth outputs results in subpar performance on LPIPS.
\romannumeral2) Single-stage diffusion-based methods encounter notable challenges with pathology images, which are characterized by intricate high-frequency textures. Among these methods, only ResShift and Resfusion perform satisfactorily.
\romannumeral3) By integrating existing methods within a two-stage restoration framework, we achieve a commendable balance between fidelity and visual appeal.
\romannumeral4) P-former excels over other methods in PSNR and SSIM, while P-diffusion demonstrates substantial improvements in LPIPS. The combination of these two approaches underscores their superior efficacy in restoring high-frequency details and enhancing overall visual quality.

Fig. \ref{fig:visualization} visualizes the images restored by different methods. 
Existing methods have various limitations in restoring pathology images. 
GRL, Restormer, and Resfusion can remove most of the defocus blur but produce over-smooth outcomes. 
ResShift produces relatively sharp results but still fails to preserve the cellular structure of the input image.
The proposed method can effectively restore the high-frequency details with excellent structural preservation, which is consistent with the quantitative results.


\textbf{Performance on 3DHistech.} 
It is worth noting that since 3DHistech does not provide multi-layer scanning data for single fields of view, we directly utilized the defocus estimator and prompt restorer pre-trained on our In-house dataset.
Despite not being extensively fine-tuned on 3DHistech, the prompt generators that we directly transferred still enabled our method to achieve SOTA performance. 
As shown in Table \ref{tab:3d_restore}, P-former excelled in distortion metrics, while the two-stage approach demonstrated superior perceptual quality.
Visualization results are provided in the supplementary material.

\begin{table}[h]
    \centering
    \resizebox{1.0\columnwidth}{!}{
    \begin{tabular}{c|cc|c}
    \hline
       \multirow{2}{*}{Method}    & \multicolumn{2}{c|}{Distortion} & \multicolumn{1}{c}{Perceptual} \\ \cline{2-4}
       & \multicolumn{1}{c}{PSNR$\uparrow$} & \multicolumn{1}{c|}{SSIM$\uparrow$}  & \multicolumn{1}{c}{LPIPS$\downarrow$} \\ \hline 

DRBNet \cite{ruan2022learning}                 & 32.83          & 0.853         & 0.131      \\
GKMNet \cite{quan2021gaussian}                 & 33.42          & 0.852         & 0.130      \\
MIMO-UNet \cite{cho2021rethinking}              & 32.40          & 0.837         & 0.169      \\
MSSNet    \cite{kim2022mssnet}              & 33.09          & 0.870         & 0.126      \\
SwinIR \cite{liang2021swinir}                 & 32.57          & 0.841         & 0.136      \\
PANet \cite{mei2023pyramid}                  & 33.24          & 0.869         & 0.129      \\
GRL \cite{li2023efficient}                    & 33.49          & 0.878         & 0.120      \\
Restormer \cite{zamir2022restormer}              & 33.46          & 0.880         & 0.125      \\
MPT + EFCR \cite{zhang2024unified}                    & 33.58          & 0.887         & 0.119      \\
Restormer + ResShift \cite{yue2024efficient}           &   31.90    & 0.865    &    0.096        \\ \hline
\textbf{P-former (ours)}                & \textbf{33.68}          & \textbf{0.903}         & 0.102      \\
\textbf{P-diffusion (ours)}             & 33.10          & 0.885         & 0.083      \\
\textbf{P-former + P-diffusion (ours)}  & 33.32          & 0.895         & \textbf{0.075}      \\ \hline
    \end{tabular}}
    \caption{Quantitative comparison on 3DHistech \cite{geng2022cervical}. The best results are highlighted in \textbf{bold}.}
    \label{tab:3d_restore}
    \end{table}

\textbf{Downstream Task on 3DHistech.} 
To further validate that our method effectively restores semantic information, we evaluated classification performance using a pre-trained model for abnormal cervical cells classification (which will be released along with our code). 
As shown in Table \ref{tab:3d_class}, our method achieved superior results across all classification metrics. 
Interestingly, we observed that classification performance correlates more with perceptual quality (Table \ref{tab:3d_restore}) than with distortion metrics.

\begin{table}[h]
\centering
\resizebox{1.0\columnwidth}{!}{
\begin{tabular}{c|cccc}
\hline
Method & AUC & Accuracy & Precision & Specificity \\ \hline
Low-quality Image & 90.90 & 82.82 & 50.35 & 83.12 \\ \hline
Restormer & 95.11 & 85.25 & 54.24 & 82.88 \\
Restormer + ResShift & 95.16 & 90.90 & 71.57 & 93.39 \\ 
\textbf{P-former } & 95.40 & 86.06 & 55.70 & 83.86 \\
\textbf{P-diffusion } & 95.83 & 91.11 & 76.24 & \textbf{95.35} \\
\textbf{P-former + P-diffusion } & \textbf{96.58} & \textbf{91.91} & \textbf{77.38} & \textbf{95.35} \\ \hline
High-quality Image  & 97.46 & 95.15 & 80.39 & 95.11 \\ \hline
\end{tabular}
}
\caption{Results of the downstream diagnostic tasks on 3DHistech. The best results are highlighted in \textbf{bold}.}
\label{tab:3d_class}
\end{table}

\subsection{Ablation Study}

\begin{table}[h]
    \centering
    \resizebox{1.0\columnwidth}{!}{
    \begin{tabular}{c|cc|c}
    \hline
       \multirow{2}{*}{Gate Controller}    & \multicolumn{2}{c|}{Distortion} & \multicolumn{1}{c}{Perceptual} \\ \cline{2-4}
       & \multicolumn{1}{c}{PSNR$\uparrow$} & \multicolumn{1}{c|}{SSIM$\uparrow$}  & \multicolumn{1}{c}{LPIPS$\downarrow$} \\ \hline 

           $F$   & \multicolumn{1}{c}{28.39$\pm$4.98} &0.8493$\pm$0.0937 & \multicolumn{1}{c}{0.1485$\pm$0.0884}  \\ 
          $P_D$   & \multicolumn{1}{c}{28.35$\pm$4.55} &0.8530$\pm$0.0894 & \multicolumn{1}{c}{0.1483$\pm$0.0847}  \\ 
           $P_{LP}$  & \multicolumn{1}{c}{28.06$\pm$3.62} &0.8610$\pm$0.0852 & \multicolumn{1}{c}{0.1357$\pm$0.0761}  \\ 
          $P_{P}$   & \multicolumn{1}{c}{28.65$\pm$5.19} &0.8579$\pm$0.0885 & \multicolumn{1}{c}{0.1297$\pm$0.0815}  \\  
          $F$+$P_D$   & \multicolumn{1}{c}{28.97$\pm$5.05} &0.8648$\pm$0.0847 & \multicolumn{1}{c}{0.1235$\pm$0.0791}  \\ 
          $F$+$P_{LP}$  & \multicolumn{1}{c}{28.98$\pm$5.07} &0.8665$\pm$0.0841 & \multicolumn{1}{c}{0.1208$\pm$0.0781}  \\  
          \textbf{$F$+$P_{P}$ (ours)}   & \multicolumn{1}{c}{\textbf{29.14$\pm$5.10}} & \textbf{0.8699$\pm$0.0828} & \multicolumn{1}{c}{0.1148$\pm$0.0746}  \\ 
          $F$+$P_D$+$P_{LP}$   & \multicolumn{1}{c}{29.01$\pm$5.11} &0.8673$\pm$0.0842 & \multicolumn{1}{c}{0.1145$\pm$0.0754}  \\ 
          $F$+$P_D$+$P_{P}$   & \multicolumn{1}{c}{29.06$\pm$5.07} & 0.8693$\pm$0.0828 & \multicolumn{1}{c}{\textbf{0.1141$\pm$0.0745}}  \\ \hline

    \end{tabular}}
    \caption{Ablation study of the use of different prompt combinations as the gate controller in P-former on the in-house dataset. $F$, $P_D$, $P_{LP}$, and $P_{P}$ refer to the input image features, defocus prompt, low-quality pathology prompt, and pathology prompt. '$+$' represents the concatenation of varied prompts. The best results are highlighted in \textbf{bold}.}
    \label{tab:ablation_stage1}
    \end{table}

\textbf{Ablation of P-former.} The essence of P-former lies in the strategic use of effective prompts as the gate controller for the MoE mechanism. 
In Table \ref{tab:ablation_stage1}, we compare the impact of various prompts and their combinations on the performance of P-former. 
The results demonstrate that the combined use of the input features ($F$) and multiple prompts consistently outperforms the individual prompts in the first four rows of Table \ref{tab:ablation_stage1}. 
Among the prompts we designed, the pathology prompt $P_{P}$ significantly improves model performance compared to the low-quality pathology prompt $P_{LP}$.
The combination of $P_{P}$ and $P_D$ did not yield significant improvements than $P_{P}$ itself, further indicating that it carries substantial information and underscoring the importance of defocus-aware prompt restoration.

\begin{table}[h]
    \centering
    \resizebox{0.9\columnwidth}{!}{
    \begin{tabular}{c|cc|c}
    \hline
        \multirow{2}{*}{Prompt}  & \multicolumn{2}{c|}{Distortion} & \multicolumn{1}{c}{Perceptual} \\ \cline{2-4}
        & \multicolumn{1}{c}{PSNR$\uparrow$} & \multicolumn{1}{c|}{SSIM$\uparrow$}  & \multicolumn{1}{c}{LPIPS$\downarrow$} \\ \hline 

         -   & \multicolumn{1}{c}{27.91$\pm$5.20} &0.8379$\pm$0.1019 & \multicolumn{1}{c}{0.0972$\pm$0.0566}  \\ 
            $P_{P}$   & \multicolumn{1}{c}{28.31$\pm$4.84} &0.8488$\pm$0.0942 & \multicolumn{1}{c}{0.0768$\pm$0.0458}  \\ 
           $P_{LE}$   & \multicolumn{1}{c}{28.04$\pm$4.62} & 0.8566$\pm$0.0869  & \multicolumn{1}{c}{0.1103$\pm$0.0689}  \\ 
            $P_{E}$   & \multicolumn{1}{c}{28.04$\pm$4.89} &0.8438$\pm$0.0952  & \multicolumn{1}{c}{0.0830$\pm$0.0478}  \\ 
            \textbf{$P_{P}$+$P_{E}$}   & \multicolumn{1}{c}{\textbf{28.91$\pm$5.18}} &\textbf{0.8578$\pm$0.0979} & \multicolumn{1}{c}{\textbf{0.0735$\pm$0.0471}}  \\ \hline 
    
    \end{tabular}}
    \caption{Ablation study of the use of different prompt combinations in P-diffusion on the in-house dataset. $P_{P}$, $P_{LE}$, and $P_{E}$ refer to the pathology prompt, low-quality edge prompt, and edge prompt. The best results are highlighted in \textbf{bold}.}
    \label{tab:ablation_stage2}
    \end{table}

\textbf{Ablation of P-diffusion.} An ablation study was conducted to assess the efficacy of the pathology prompt $P_{P}$ and edge prompt $P_{E}$ in P-diffusion.
Experiments in Table \ref{tab:ablation_stage2} take $I^{'}$ as the input, illustrating that both prompts contribute to enhancing performance during the fine stage. 
The integration of edge fusion notably elevates distortion metrics, with the defocus prompt $P_D$ emerging as pivotal for estimating the confidence of the low-quality edge prompt $P_{LE}$.
The model with both prompts attains peak performance, underscoring their ability to effectively craft realistic patterns tailored to diverse pathology image types while preserving input image structure.

\subsection{Effects of Prompt Generation}

\begin{figure}[h]
    \centering
    \includegraphics[width=1\linewidth]{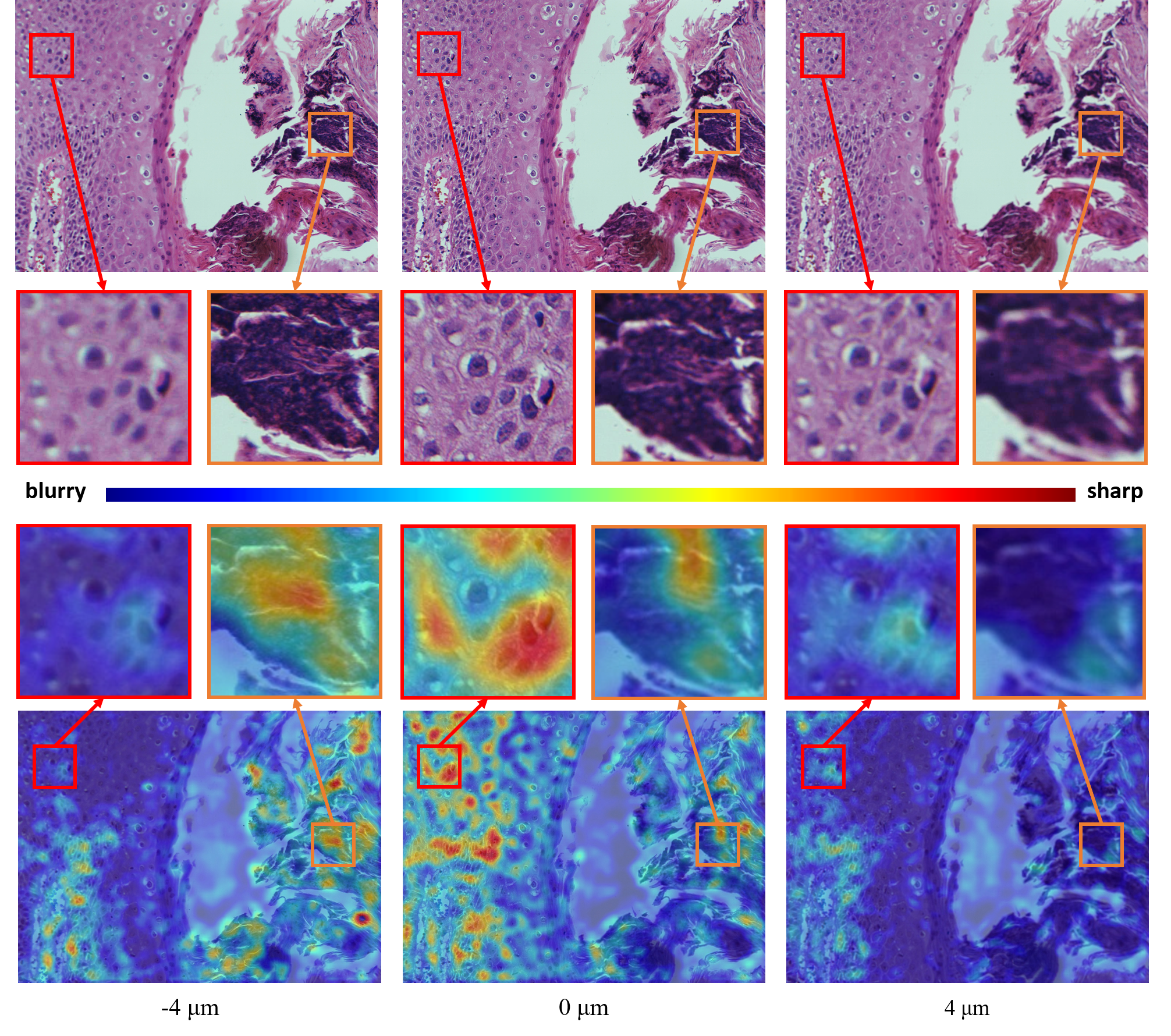}
    \caption{Within the same field of view, the original image and the heatmap generated by the defocus estimator exhibit variations with changes in the focal distance. The red and orange boxes highlight localized magnified views for closer inspection.}
    \label{fig:defocus_heatmap}
\end{figure}

\textbf{Effects of Defocus Estimator $D_{\phi }$.}
Figure \ref{fig:defocus_heatmap} provides an intuitive demonstration of the changes induced by focal distance. 
$P_D$ is passed through the linear layer that predicts the distance to reduce the number of channels to 1, then interpolated and taken absolute value, producing the heatmap to reflect the local degree of blur.
We can observe that the tissue within the red box becomes clearer and then blurrier as the focal distance varies, while the content within the orange box consistently becomes more blurred. 
This phenomenon indicates that even within a unified field of view, the degree of blur can differ due to varying tissue thicknesses. 
Our trained defocus estimator accurately captures these differences, which can be leveraged to enhance the image restoration process.
More details of $D_{\phi }$ are provided in the supplementary material.

\begin{figure}[h]
    \centering
    \includegraphics[width=1\linewidth]{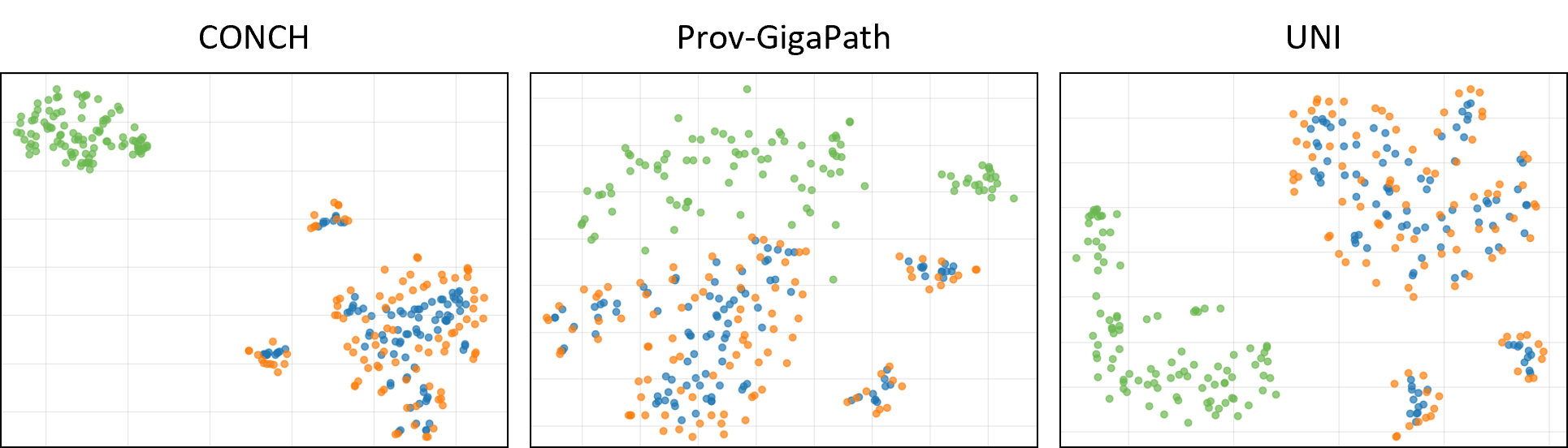}
    \caption{The visualization obtained using t-SNE illustrates the effect of defocus-aware prompt restoration, showcasing the feature distributions of \textcolor[HTML]{6ab84f}{$P_{LP}$ (green)}, \textcolor[HTML]{1f77b4}{$P_{P}$ (blue)} and \textcolor[HTML]{ff7f0e}{ $P_{HP}$ (orange)}.}
    \label{fig:restore_prompt}
\end{figure}

\textbf{Effects of Restoring Pathology Prompt $P_P$.}
To provide more intuitive evidence that $P_P$ is a valid prior, we utilized t-SNE to visualize $P_{LP}$, $P_{P}$, and $P_{HP}$ for various pathology foundation models. 
The results, as illustrated in Figure \ref{fig:restore_prompt}, clearly demonstrate that the distribution of $P_{P}$ is significantly closer to that of $P_{HP}$ compared to $P_{LP}$. This observation indicates that the pathology prompt has been effectively rectified. 
Numerical results are provided in the supplementary material.

\section{Conclusion}
\label{sec:conclusion}

This paper introduces a novel approach that leverages a mixture of prompts to address and solve the key challenges in pathology image restoration. Furthermore, it presents the design of P-former and P-diffusion to optimize the application of these prompts.
Our method attains cutting-edge performance on distortion and perceptual metrics. 
Additionally, we contributed a set of diagnostic annotations to the public dataset, further demonstrating the effectiveness of our method.
However, one limitation of our study is the diffusion process, which slows down the restoration speed.
Consequently, future work needs to focus on accelerating the diffusion process to overcome this challenge.

{
    \small
    \bibliographystyle{ieeenat_fullname}
    \bibliography{main}

\begin{thebibliography}{64}
\providecommand{\natexlab}[1]{#1}
\providecommand{\url}[1]{\texttt{#1}}
\expandafter\ifx\csname urlstyle\endcsname\relax
  \providecommand{\doi}[1]{doi: #1}\else
  \providecommand{\doi}{doi: \begingroup \urlstyle{rm}\Url}\fi

\bibitem[Bhat and Koundal(2021)]{bhat2021multi}
Shiveta Bhat and Deepika Koundal.
\newblock Multi-focus image fusion techniques: a survey.
\newblock \emph{Artificial Intelligence Review}, 54\penalty0 (8):\penalty0 5735--5787, 2021.

\bibitem[Bian et~al.(2020)Bian, Guo, Jiang, Zhu, Wang, Song, Zhang, Hoshino, and Zheng]{bian2020autofocusing}
Zichao Bian, Chengfei Guo, Shaowei Jiang, Jiakai Zhu, Ruihai Wang, Pengming Song, Zibang Zhang, Kazunori Hoshino, and Guoan Zheng.
\newblock Autofocusing technologies for whole slide imaging and automated microscopy.
\newblock \emph{Journal of Biophotonics}, 13\penalty0 (12):\penalty0 e202000227, 2020.

\bibitem[Bulten et~al.(2022)Bulten, Kartasalo, Chen, Str{\"o}m, Pinckaers, Nagpal, Cai, Steiner, Van~Boven, Vink, et~al.]{bulten2022artificial}
Wouter Bulten, Kimmo Kartasalo, Po-Hsuan~Cameron Chen, Peter Str{\"o}m, Hans Pinckaers, Kunal Nagpal, Yuannan Cai, David~F Steiner, Hester Van~Boven, Robert Vink, et~al.
\newblock Artificial intelligence for diagnosis and gleason grading of prostate cancer: the panda challenge.
\newblock \emph{Nature medicine}, 28\penalty0 (1):\penalty0 154--163, 2022.

\bibitem[Chen et~al.(2024)Chen, Ding, Lu, Williamson, Jaume, Song, Chen, Zhang, Shao, Shaban, et~al.]{chen2024towards}
Richard~J Chen, Tong Ding, Ming~Y Lu, Drew~FK Williamson, Guillaume Jaume, Andrew~H Song, Bowen Chen, Andrew Zhang, Daniel Shao, Muhammad Shaban, et~al.
\newblock Towards a general-purpose foundation model for computational pathology.
\newblock \emph{Nature Medicine}, 30\penalty0 (3):\penalty0 850--862, 2024.

\bibitem[Chen et~al.(2023)Chen, Zhang, Gu, Yuan, Kong, Chen, and Yang]{chen2023image}
Zheng Chen, Yulun Zhang, Jinjin Gu, Xin Yuan, Linghe Kong, Guihai Chen, and Xiaokang Yang.
\newblock Image super-resolution with text prompt diffusion.
\newblock \emph{arXiv preprint arXiv:2311.14282}, 2023.

\bibitem[Cho et~al.(2021)Cho, Ji, Hong, Jung, and Ko]{cho2021rethinking}
Sung-Jin Cho, Seo-Won Ji, Jun-Pyo Hong, Seung-Won Jung, and Sung-Jea Ko.
\newblock Rethinking coarse-to-fine approach in single image deblurring.
\newblock In \emph{Proceedings of the IEEE/CVF international conference on computer vision}, pages 4641--4650, 2021.

\bibitem[Deng et~al.(2009)Deng, Dong, Socher, Li, Li, and Fei-Fei]{deng2009imagenet}
Jia Deng, Wei Dong, Richard Socher, Li-Jia Li, Kai Li, and Li Fei-Fei.
\newblock Imagenet: A large-scale hierarchical image database.
\newblock In \emph{2009 IEEE conference on computer vision and pattern recognition}, pages 248--255. Ieee, 2009.

\bibitem[et~al(2022)]{geng2022cervical}
Geng et al.
\newblock Cervical cytopathology image refocusing via multi-scale attention features and domain normalization.
\newblock \emph{Medical Image Analysis}, 2022.

\bibitem[et~al(2024)]{yang2024rat}
Yang et al.
\newblock Region attention transformer for medical image restoration.
\newblock In \emph{MICCAI}, 2024.

\bibitem[Guo et~al.(2024)Guo, Gao, Lu, Zhu, Liu, and He]{guo2024onerestore}
Yu Guo, Yuan Gao, Yuxu Lu, Huilin Zhu, Ryan~Wen Liu, and Shengfeng He.
\newblock Onerestore: A universal restoration framework for composite degradation.
\newblock \emph{arXiv preprint arXiv:2407.04621}, 2024.

\bibitem[He et~al.(2016)He, Zhang, Ren, and Sun]{he2016deep}
Kaiming He, Xiangyu Zhang, Shaoqing Ren, and Jian Sun.
\newblock Deep residual learning for image recognition.
\newblock In \emph{Proceedings of the IEEE conference on computer vision and pattern recognition}, pages 770--778, 2016.

\bibitem[He et~al.(2020)He, Fan, Wu, Xie, and Girshick]{he2020momentum}
Kaiming He, Haoqi Fan, Yuxin Wu, Saining Xie, and Ross Girshick.
\newblock Momentum contrast for unsupervised visual representation learning.
\newblock In \emph{Proceedings of the IEEE/CVF conference on computer vision and pattern recognition}, pages 9729--9738, 2020.

\bibitem[He et~al.(2022)He, Chen, Xie, Li, Doll{\'a}r, and Girshick]{he2022masked}
Kaiming He, Xinlei Chen, Saining Xie, Yanghao Li, Piotr Doll{\'a}r, and Ross Girshick.
\newblock Masked autoencoders are scalable vision learners.
\newblock In \emph{Proceedings of the IEEE/CVF conference on computer vision and pattern recognition}, pages 16000--16009, 2022.

\bibitem[Huang et~al.(2023)Huang, Bianchi, Yuksekgonul, Montine, and Zou]{huang2023visual}
Zhi Huang, Federico Bianchi, Mert Yuksekgonul, Thomas~J Montine, and James Zou.
\newblock A visual--language foundation model for pathology image analysis using medical twitter.
\newblock \emph{Nature medicine}, 29\penalty0 (9):\penalty0 2307--2316, 2023.

\bibitem[Jacobs et~al.(1991)Jacobs, Jordan, Nowlan, and Hinton]{jacobs1991adaptive}
Robert~A Jacobs, Michael~I Jordan, Steven~J Nowlan, and Geoffrey~E Hinton.
\newblock Adaptive mixtures of local experts.
\newblock \emph{Neural computation}, 3\penalty0 (1):\penalty0 79--87, 1991.

\bibitem[Kang et~al.(2023)Kang, Song, Park, Yoo, and Pereira]{kang2023benchmarking}
Mingu Kang, Heon Song, Seonwook Park, Donggeun Yoo, and S{\'e}rgio Pereira.
\newblock Benchmarking self-supervised learning on diverse pathology datasets.
\newblock In \emph{Proceedings of the IEEE/CVF Conference on Computer Vision and Pattern Recognition}, pages 3344--3354, 2023.

\bibitem[Kim et~al.(2022)Kim, Lee, and Cho]{kim2022mssnet}
Kiyeon Kim, Seungyong Lee, and Sunghyun Cho.
\newblock Mssnet: Multi-scale-stage network for single image deblurring.
\newblock In \emph{European conference on computer vision}, pages 524--539. Springer, 2022.

\bibitem[Kirillov et~al.(2023)Kirillov, Mintun, Ravi, Mao, Rolland, Gustafson, Xiao, Whitehead, Berg, Lo, et~al.]{kirillov2023segment}
Alexander Kirillov, Eric Mintun, Nikhila Ravi, Hanzi Mao, Chloe Rolland, Laura Gustafson, Tete Xiao, Spencer Whitehead, Alexander~C Berg, Wan-Yen Lo, et~al.
\newblock Segment anything.
\newblock In \emph{Proceedings of the IEEE/CVF international conference on computer vision}, pages 4015--4026, 2023.

\bibitem[Kirkland(1998)]{kirkland1998advanced}
Earl~J Kirkland.
\newblock \emph{Advanced computing in electron microscopy}.
\newblock Springer, 1998.

\bibitem[Ledig et~al.(2017)Ledig, Theis, Husz{\'a}r, Caballero, Cunningham, Acosta, Aitken, Tejani, Totz, Wang, et~al.]{ledig2017photo}
Christian Ledig, Lucas Theis, Ferenc Husz{\'a}r, Jose Caballero, Andrew Cunningham, Alejandro Acosta, Andrew Aitken, Alykhan Tejani, Johannes Totz, Zehan Wang, et~al.
\newblock Photo-realistic single image super-resolution using a generative adversarial network.
\newblock In \emph{Proceedings of the IEEE conference on computer vision and pattern recognition}, pages 4681--4690, 2017.

\bibitem[Li et~al.(2024)Li, Li, Lu, Feng, Guo, Zhao, Zhang, and Chen]{li2024promptcir}
Bingchen Li, Xin Li, Yiting Lu, Ruoyu Feng, Mengxi Guo, Shijie Zhao, Li Zhang, and Zhibo Chen.
\newblock Promptcir: Blind compressed image restoration with prompt learning.
\newblock \emph{arXiv preprint arXiv:2404.17433}, 2024.

\bibitem[Li et~al.(2023)Li, Fan, Xiang, Demandolx, Ranjan, Timofte, and Van~Gool]{li2023efficient}
Yawei Li, Yuchen Fan, Xiaoyu Xiang, Denis Demandolx, Rakesh Ranjan, Radu Timofte, and Luc Van~Gool.
\newblock Efficient and explicit modelling of image hierarchies for image restoration.
\newblock In \emph{Proceedings of the IEEE/CVF Conference on Computer Vision and Pattern Recognition}, pages 18278--18289, 2023.

\bibitem[Liang et~al.(2021)Liang, Cao, Sun, Zhang, Van~Gool, and Timofte]{liang2021swinir}
Jingyun Liang, Jiezhang Cao, Guolei Sun, Kai Zhang, Luc Van~Gool, and Radu Timofte.
\newblock Swinir: Image restoration using swin transformer.
\newblock In \emph{Proceedings of the IEEE/CVF international conference on computer vision}, pages 1833--1844, 2021.

\bibitem[Lin et~al.(2024)Lin, Zhang, Wei, Ren, Jiang, Tian, and Zuo]{lin2024improving}
Jingbo Lin, Zhilu Zhang, Yuxiang Wei, Dongwei Ren, Dongsheng Jiang, Qi Tian, and Wangmeng Zuo.
\newblock Improving image restoration through removing degradations in textual representations.
\newblock In \emph{Proceedings of the IEEE/CVF Conference on Computer Vision and Pattern Recognition}, pages 2866--2878, 2024.

\bibitem[Lin et~al.(2014)Lin, Maire, Belongie, Hays, Perona, Ramanan, Doll{\'a}r, and Zitnick]{lin2014microsoft}
Tsung-Yi Lin, Michael Maire, Serge Belongie, James Hays, Pietro Perona, Deva Ramanan, Piotr Doll{\'a}r, and C~Lawrence Zitnick.
\newblock Microsoft coco: Common objects in context.
\newblock In \emph{Computer Vision--ECCV 2014: 13th European Conference, Zurich, Switzerland, September 6-12, 2014, Proceedings, Part V 13}, pages 740--755. Springer, 2014.

\bibitem[Lin et~al.(2023)Lin, He, Chen, Lyu, Dai, Yu, Ouyang, Qiao, and Dong]{lin2023diffbir}
Xinqi Lin, Jingwen He, Ziyan Chen, Zhaoyang Lyu, Bo Dai, Fanghua Yu, Wanli Ouyang, Yu Qiao, and Chao Dong.
\newblock Diffbir: Towards blind image restoration with generative diffusion prior.
\newblock \emph{arXiv preprint arXiv:2308.15070}, 2023.

\bibitem[Liu et~al.(2024{\natexlab{a}})Liu, Wang, Fan, Wang, Tang, and Qu]{liu2024residual}
Jiawei Liu, Qiang Wang, Huijie Fan, Yinong Wang, Yandong Tang, and Liangqiong Qu.
\newblock Residual denoising diffusion models.
\newblock In \emph{Proceedings of the IEEE/CVF Conference on Computer Vision and Pattern Recognition}, pages 2773--2783, 2024{\natexlab{a}}.

\bibitem[Liu et~al.(2020)Liu, Wang, Cheng, Li, and Chen]{liu2020multi}
Yu Liu, Lei Wang, Juan Cheng, Chang Li, and Xun Chen.
\newblock Multi-focus image fusion: A survey of the state of the art.
\newblock \emph{Information Fusion}, 64:\penalty0 71--91, 2020.

\bibitem[Liu et~al.(2024{\natexlab{b}})Liu, Ke, Liu, Zhao, and Lau]{liu2024diff}
Yuhao Liu, Zhanghan Ke, Fang Liu, Nanxuan Zhao, and Rynson~WH Lau.
\newblock Diff-plugin: Revitalizing details for diffusion-based low-level tasks.
\newblock In \emph{Proceedings of the IEEE/CVF Conference on Computer Vision and Pattern Recognition}, pages 4197--4208, 2024{\natexlab{b}}.

\bibitem[Lu et~al.(2024{\natexlab{a}})Lu, Chen, Williamson, Chen, Liang, Ding, Jaume, Odintsov, Le, Gerber, et~al.]{lu2024avisionlanguage}
Ming~Y Lu, Bowen Chen, Drew~FK Williamson, Richard~J Chen, Ivy Liang, Tong Ding, Guillaume Jaume, Igor Odintsov, Long~Phi Le, Georg Gerber, et~al.
\newblock A visual-language foundation model for computational pathology.
\newblock \emph{Nature Medicine}, 30:\penalty0 863–874, 2024{\natexlab{a}}.

\bibitem[Lu et~al.(2024{\natexlab{b}})Lu, Chen, Williamson, Chen, Liang, Ding, Jaume, Odintsov, Le, Gerber, et~al.]{lu2024visual}
Ming~Y Lu, Bowen Chen, Drew~FK Williamson, Richard~J Chen, Ivy Liang, Tong Ding, Guillaume Jaume, Igor Odintsov, Long~Phi Le, Georg Gerber, et~al.
\newblock A visual-language foundation model for computational pathology.
\newblock \emph{Nature Medicine}, 30\penalty0 (3):\penalty0 863--874, 2024{\natexlab{b}}.

\bibitem[Luo et~al.(2023{\natexlab{a}})Luo, Gustafsson, Zhao, Sj{\"o}lund, and Sch{\"o}n]{luo2023controlling}
Ziwei Luo, Fredrik~K Gustafsson, Zheng Zhao, Jens Sj{\"o}lund, and Thomas~B Sch{\"o}n.
\newblock Controlling vision-language models for universal image restoration.
\newblock \emph{arXiv preprint arXiv:2310.01018}, 3\penalty0 (8), 2023{\natexlab{a}}.

\bibitem[Luo et~al.(2023{\natexlab{b}})Luo, Gustafsson, Zhao, Sj{\"o}lund, and Sch{\"o}n]{luo2023image}
Ziwei Luo, Fredrik~K Gustafsson, Zheng Zhao, Jens Sj{\"o}lund, and Thomas~B Sch{\"o}n.
\newblock Image restoration with mean-reverting stochastic differential equations.
\newblock \emph{arXiv preprint arXiv:2301.11699}, 2023{\natexlab{b}}.

\bibitem[Ma et~al.(2022)Ma, Rathgeb, Mubarak, Tran, and Fei]{ma2022unsupervised}
Ling Ma, Armand Rathgeb, Hasan Mubarak, Minh Tran, and Baowei Fei.
\newblock Unsupervised super-resolution reconstruction of hyperspectral histology images for whole-slide imaging.
\newblock \emph{Journal of biomedical optics}, 27\penalty0 (5):\penalty0 056502--056502, 2022.

\bibitem[Marcelino(2018)]{marcelino2018transfer}
Pedro Marcelino.
\newblock Transfer learning from pre-trained models.
\newblock \emph{Towards data science}, 10\penalty0 (330):\penalty0 23, 2018.

\bibitem[Martin and Thomas(1995)]{martin1995experimental}
David~C Martin and Edwin~L Thomas.
\newblock Experimental high-resolution electron microscopy of polymers.
\newblock \emph{Polymer}, 36\penalty0 (9):\penalty0 1743--1759, 1995.

\bibitem[Mei et~al.(2023)Mei, Fan, Zhang, Yu, Zhou, Liu, Fu, Huang, and Shi]{mei2023pyramid}
Yiqun Mei, Yuchen Fan, Yulun Zhang, Jiahui Yu, Yuqian Zhou, Ding Liu, Yun Fu, Thomas~S Huang, and Humphrey Shi.
\newblock Pyramid attention network for image restoration.
\newblock \emph{International Journal of Computer Vision}, 131\penalty0 (12):\penalty0 3207--3225, 2023.

\bibitem[Nechaev et~al.(2024)Nechaev, Pchelnikov, and Ivanova]{nechaev2024hibou}
Dmitry Nechaev, Alexey Pchelnikov, and Ekaterina Ivanova.
\newblock Hibou: A family of foundational vision transformers for pathology.
\newblock \emph{arXiv preprint arXiv:2406.05074}, 2024.

\bibitem[Potlapalli et~al.(2024)Potlapalli, Zamir, Khan, and Shahbaz~Khan]{potlapalli2024promptir}
Vaishnav Potlapalli, Syed~Waqas Zamir, Salman~H Khan, and Fahad Shahbaz~Khan.
\newblock Promptir: Prompting for all-in-one image restoration.
\newblock \emph{Advances in Neural Information Processing Systems}, 36, 2024.

\bibitem[Qi et~al.(2023)Qi, Tu, Ye, Delbracio, Milanfar, Chen, and Talebi]{qi2023tip}
Chenyang Qi, Zhengzhong Tu, Keren Ye, Mauricio Delbracio, Peyman Milanfar, Qifeng Chen, and Hossein Talebi.
\newblock Tip: Text-driven image processing with semantic and restoration instructions.
\newblock \emph{arXiv preprint arXiv:2312.11595}, 2023.

\bibitem[Quan et~al.(2021)Quan, Wu, and Ji]{quan2021gaussian}
Yuhui Quan, Zicong Wu, and Hui Ji.
\newblock Gaussian kernel mixture network for single image defocus deblurring.
\newblock \emph{Advances in Neural Information Processing Systems}, 34:\penalty0 20812--20824, 2021.

\bibitem[Quan et~al.(2023{\natexlab{a}})Quan, Wu, and Ji]{quan2023neumann}
Yuhui Quan, Zicong Wu, and Hui Ji.
\newblock Neumann network with recursive kernels for single image defocus deblurring.
\newblock In \emph{Proceedings of the IEEE/CVF Conference on Computer Vision and Pattern Recognition}, pages 5754--5763, 2023{\natexlab{a}}.

\bibitem[Quan et~al.(2023{\natexlab{b}})Quan, Yao, and Ji]{quan2023single}
Yuhui Quan, Xin Yao, and Hui Ji.
\newblock Single image defocus deblurring via implicit neural inverse kernels.
\newblock In \emph{Proceedings of the IEEE/CVF International Conference on Computer Vision}, pages 12600--12610, 2023{\natexlab{b}}.

\bibitem[Reimer(2013)]{reimer2013transmission}
Ludwig Reimer.
\newblock \emph{Transmission electron microscopy: physics of image formation and microanalysis}.
\newblock Springer, 2013.

\bibitem[Rombach et~al.(2022)Rombach, Blattmann, Lorenz, Esser, and Ommer]{rombach2022high}
Robin Rombach, Andreas Blattmann, Dominik Lorenz, Patrick Esser, and Bj{\"o}rn Ommer.
\newblock High-resolution image synthesis with latent diffusion models.
\newblock In \emph{Proceedings of the IEEE/CVF conference on computer vision and pattern recognition}, pages 10684--10695, 2022.

\bibitem[Rong et~al.(2023)Rong, Wang, Zhang, Wen, Cheng, Jia, Yang, Xie, Zhan, and Xiao]{rong2023enhanced}
Ruichen Rong, Shidan Wang, Xinyi Zhang, Zhuoyu Wen, Xian Cheng, Liwei Jia, Donghan~M Yang, Yang Xie, Xiaowei Zhan, and Guanghua Xiao.
\newblock Enhanced pathology image quality with restore--generative adversarial network.
\newblock \emph{The American Journal of Pathology}, 193\penalty0 (4):\penalty0 404--416, 2023.

\bibitem[Ronneberger et~al.(2015)Ronneberger, Fischer, and Brox]{ronneberger2015u}
Olaf Ronneberger, Philipp Fischer, and Thomas Brox.
\newblock U-net: Convolutional networks for biomedical image segmentation.
\newblock In \emph{Medical image computing and computer-assisted intervention--MICCAI 2015: 18th international conference, Munich, Germany, October 5-9, 2015, proceedings, part III 18}, pages 234--241. Springer, 2015.

\bibitem[Ruan et~al.(2022)Ruan, Chen, Li, and Lam]{ruan2022learning}
Lingyan Ruan, Bin Chen, Jizhou Li, and Miuling Lam.
\newblock Learning to deblur using light field generated and real defocus images.
\newblock In \emph{Proceedings of the IEEE/CVF Conference on Computer Vision and Pattern Recognition}, pages 16304--16313, 2022.

\bibitem[Saillard et~al.(2024)Saillard, Jenatton, Llinares-López, Mariet, Cahané, Durand, and Vert]{hoptimus0}
Charlie Saillard, Rodolphe Jenatton, Felipe Llinares-López, Zelda Mariet, David Cahané, Eric Durand, and Jean-Philippe Vert.
\newblock H-optimus-0, 2024.

\bibitem[Shazeer et~al.(2017)Shazeer, Mirhoseini, Maziarz, Davis, Le, Hinton, and Dean]{shazeer2017outrageously}
Noam Shazeer, Azalia Mirhoseini, Krzysztof Maziarz, Andy Davis, Quoc Le, Geoffrey Hinton, and Jeff Dean.
\newblock Outrageously large neural networks: The sparsely-gated mixture-of-experts layer.
\newblock \emph{arXiv preprint arXiv:1701.06538}, 2017.

\bibitem[Shen et~al.(2022)Shen, Luo, Shen, and Ke]{shen2022randstainna}
Yiqing Shen, Yulin Luo, Dinggang Shen, and Jing Ke.
\newblock Randstainna: Learning stain-agnostic features from histology slides by bridging stain augmentation and normalization.
\newblock In \emph{International Conference on Medical Image Computing and Computer-Assisted Intervention}, pages 212--221. Springer, 2022.

\bibitem[Wade(1992)]{wade1992brief}
RH Wade.
\newblock A brief look at imaging and contrast transfer.
\newblock \emph{Ultramicroscopy}, 46\penalty0 (1-4):\penalty0 145--156, 1992.

\bibitem[Wang et~al.(2022)Wang, Cun, Bao, Zhou, Liu, and Li]{wang2022uformer}
Zhendong Wang, Xiaodong Cun, Jianmin Bao, Wengang Zhou, Jianzhuang Liu, and Houqiang Li.
\newblock Uformer: A general u-shaped transformer for image restoration.
\newblock In \emph{Proceedings of the IEEE/CVF conference on computer vision and pattern recognition}, pages 17683--17693, 2022.

\bibitem[Wright et~al.(2020)Wright, Dunn, Hale, Hutchins, and Treanor]{wright2020effect}
Alexander~I Wright, Catriona~M Dunn, Michael Hale, Gordon~GA Hutchins, and Darren~E Treanor.
\newblock The effect of quality control on accuracy of digital pathology image analysis.
\newblock \emph{IEEE Journal of Biomedical and Health Informatics}, 25\penalty0 (2):\penalty0 307--314, 2020.

\bibitem[Xia et~al.(2023)Xia, Zhang, Wang, Wang, Wu, Tian, Yang, and Van~Gool]{xia2023diffir}
Bin Xia, Yulun Zhang, Shiyin Wang, Yitong Wang, Xinglong Wu, Yapeng Tian, Wenming Yang, and Luc Van~Gool.
\newblock Diffir: Efficient diffusion model for image restoration.
\newblock In \emph{Proceedings of the IEEE/CVF International Conference on Computer Vision}, pages 13095--13105, 2023.

\bibitem[Xu et~al.(2024{\natexlab{a}})Xu, Usuyama, Bagga, Zhang, Rao, Naumann, Wong, Gero, Gonz{\'a}lez, Gu, et~al.]{xu2024whole}
Hanwen Xu, Naoto Usuyama, Jaspreet Bagga, Sheng Zhang, Rajesh Rao, Tristan Naumann, Cliff Wong, Zelalem Gero, Javier Gonz{\'a}lez, Yu Gu, et~al.
\newblock A whole-slide foundation model for digital pathology from real-world data.
\newblock \emph{Nature}, pages 1--8, 2024{\natexlab{a}}.

\bibitem[Xu et~al.(2024{\natexlab{b}})Xu, Kong, Hu, Liu, and Bao]{xu2024boosting}
Xiaogang Xu, Shu Kong, Tao Hu, Zhe Liu, and Hujun Bao.
\newblock Boosting image restoration via priors from pre-trained models.
\newblock In \emph{Proceedings of the IEEE/CVF Conference on Computer Vision and Pattern Recognition}, pages 2900--2909, 2024{\natexlab{b}}.

\bibitem[Yue et~al.(2024)Yue, Wang, and Loy]{yue2024efficient}
Zongsheng Yue, Jianyi Wang, and Chen~Change Loy.
\newblock Efficient diffusion model for image restoration by residual shifting.
\newblock \emph{arXiv preprint arXiv:2403.07319}, 2024.

\bibitem[Zamir et~al.(2022)Zamir, Arora, Khan, Hayat, Khan, and Yang]{zamir2022restormer}
Syed~Waqas Zamir, Aditya Arora, Salman Khan, Munawar Hayat, Fahad~Shahbaz Khan, and Ming-Hsuan Yang.
\newblock Restormer: Efficient transformer for high-resolution image restoration.
\newblock In \emph{Proceedings of the IEEE/CVF conference on computer vision and pattern recognition}, pages 5728--5739, 2022.

\bibitem[Zhang et~al.(2024)Zhang, Zheng, Yan, Fang, and Cheng]{zhang2024unified}
Yuelin Zhang, Pengyu Zheng, Wanquan Yan, Chengyu Fang, and Shing~Shin Cheng.
\newblock A unified framework for microscopy defocus deblur with multi-pyramid transformer and contrastive learning.
\newblock In \emph{Proceedings of the IEEE/CVF Conference on Computer Vision and Pattern Recognition}, pages 11125--11136, 2024.

\bibitem[Zhao et~al.(2023)Zhao, Liu, Wu, Wang, Li, Wang, Teng, Liu, Cui, Wang, et~al.]{zhao2023clip}
Zihao Zhao, Yuxiao Liu, Han Wu, Mei Wang, Yonghao Li, Sheng Wang, Lin Teng, Disheng Liu, Zhiming Cui, Qian Wang, et~al.
\newblock Clip in medical imaging: A comprehensive survey.
\newblock \emph{arXiv preprint arXiv:2312.07353}, 2023.

\bibitem[Zhenning et~al.(2023)Zhenning, Changsheng, Bin, Xueshuo, Along, Qiaoying, and Tao]{zhenning2023resfusion}
Shi Zhenning, Dong Changsheng, Pan Bin, Xie Xueshuo, He Along, Qu Qiaoying, and Li Tao.
\newblock Resfusion: Prior residual noise embedded denoising diffusion probabilistic models.
\newblock \emph{arXiv preprint arXiv:2311.14900}, 2023.

\bibitem[Zhou et~al.(2022)Zhou, Yu, Luo, Wang, and Li]{zhou2022mimco}
Qiang Zhou, Chaohui Yu, Hao Luo, Zhibin Wang, and Hao Li.
\newblock Mimco: Masked image modeling pre-training with contrastive teacher.
\newblock In \emph{Proceedings of the 30th ACM International Conference on Multimedia}, pages 4487--4495, 2022.

\bibitem[Zhuang et~al.(2020)Zhuang, Qi, Duan, Xi, Zhu, Zhu, Xiong, and He]{zhuang2020comprehensive}
Fuzhen Zhuang, Zhiyuan Qi, Keyu Duan, Dongbo Xi, Yongchun Zhu, Hengshu Zhu, Hui Xiong, and Qing He.
\newblock A comprehensive survey on transfer learning.
\newblock \emph{Proceedings of the IEEE}, 109\penalty0 (1):\penalty0 43--76, 2020.

\end{thebibliography}
}

\clearpage
\setcounter{page}{1}
\maketitlesupplementary
\section{Supplementary Experiments}

\begin{table}[h]
    \centering
    \resizebox{1.0\columnwidth}{!}{
    \begin{tabular}{c|ccc}
    \hline
       \multicolumn{1}{c|}{Target} & \multicolumn{1}{c}{MAE(CTF)$\downarrow$} 
       & \multicolumn{1}{c}{MAE(Distance)$\downarrow$} & \multicolumn{1}{c}{Z-accuracy$\uparrow$}   \\ \hline 

           CTF   & \multicolumn{1}{c}{\textbf{0.0523}} & N/A & N/A  \\ 
        Distance   & N/A & 0.7106 & \multicolumn{1}{c}{0.5936}  \\ 
           CTF+Distance   & \multicolumn{1}{c}{0.0610} &\textbf{0.6517} & \multicolumn{1}{c}{\textbf{0.6758}}  \\ \hline
    \end{tabular}}
    \caption{Ablation study of training defocus encoder. The best results are highlighted in \textbf{bold}.}
    \label{tab:dp_training}
    \end{table}

\textbf{Multi-task Learning for Defocus Estimator $D_{\phi }$.} Table \ref{tab:dp_training} demonstrates the effects of training the defocus encoder with different objectives. 
It is evident that incorporating the CTF prediction task enhances the model's performance on MAE and Z-accuracy. 
Although the predictive performance for CTF shows a slight decrease compared to standalone prediction, considering that focus estimation more accurately reflects the degree of image blur, we opted for a multi-task learning approach.

\begin{table}[h]
    \centering
    \resizebox{1.0\columnwidth}{!}{
    \begin{tabular}{c|cc|c}
    \hline
       \multirow{2}{*}{Encoder}    & \multicolumn{2}{c|}{Distortion} & \multicolumn{1}{c}{Perceptual} \\ \cline{2-4}
       & \multicolumn{1}{c}{PSNR$\uparrow$} & \multicolumn{1}{c|}{SSIM$\uparrow$}  & \multicolumn{1}{c}{LPIPS$\downarrow$} \\ \hline 

           UNI \cite{chen2024towards}   & \multicolumn{1}{c}{28.23$\pm$4.62} &0.8486$\pm$0.0938 & \multicolumn{1}{c}{0.0768$\pm$0.0450}  \\ 
          CONCH \cite{lu2024avisionlanguage}  & \multicolumn{1}{c}{28.28$\pm$4.84} &0.8480$\pm$0.0940 & \multicolumn{1}{c}{\textbf{0.0765$\pm$0.0456}}  \\  
          Prov-GigaPath \cite{xu2024whole}   & \multicolumn{1}{c}{\textbf{28.31$\pm$4.84}} &\textbf{0.8488$\pm$0.0942} & \multicolumn{1}{c}{0.0768$\pm$0.0458}  \\ \hline 
    \end{tabular}}
    \caption{Comparison of the pathology prompt extracted from different pathology foundation models. The baseline utilizes Restormer as the coarse stage, and ResShift as the fine stage. The best results are highlighted in \textbf{bold}.}
    \label{tab:encoder_ablation}
    \end{table}

\textbf{Pathology Foundation Model Selection.} To ensure a fair comparison among different pathological foundation models, we freeze the image encoders for prompt generation and perform fine-tuning to the parameters of P-diffusion, which is enhanced solely by the pathology prompt $P_P$. From the results in Table \ref{tab:encoder_ablation}, we observe that prompts extracted from various pathology-based models can all enhance the diffusion process. 
but did not exhibit significant differences in performance across our dataset.

\begin{figure}[h]
    \centering
    \includegraphics[width=1\linewidth]{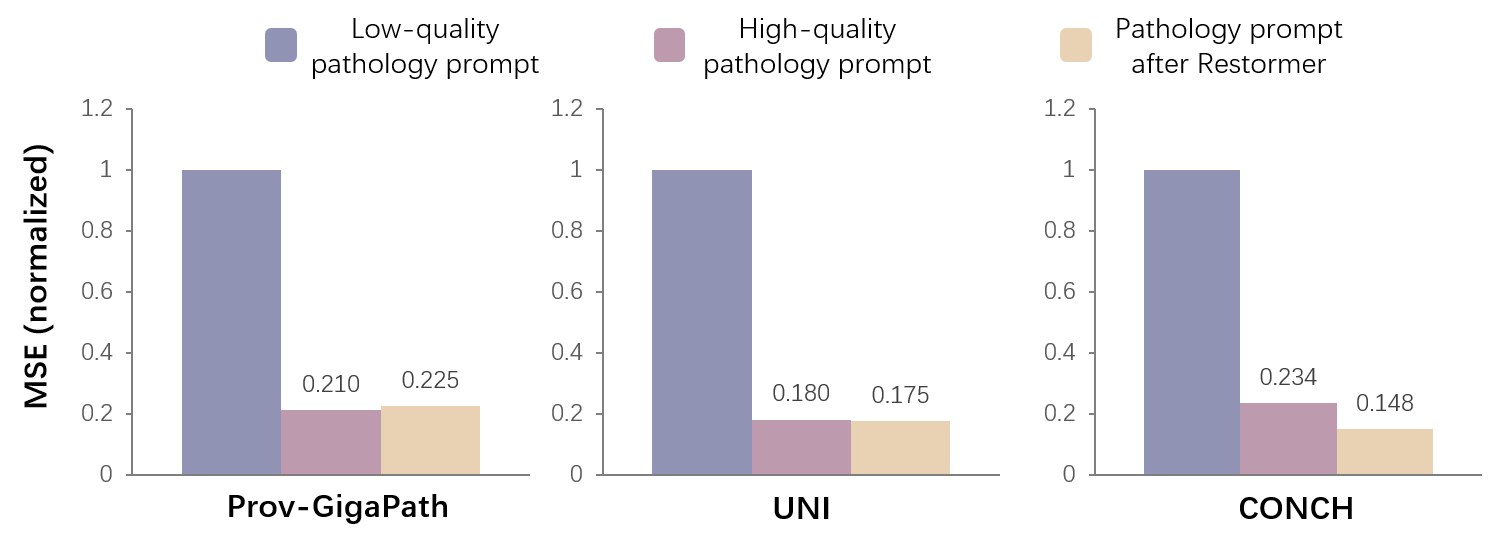}
    \caption{Effects of defocus-aware prompt restoration. Mean Squared Error (MSE) measures the distance between various prompts and the high-quality pathology prompt $P_{HP}$.}
    \label{fig:restore_prompt_mse}
\end{figure}

\textbf{Effects of Restoring Pathology Prompt $P_P$.}
To demonstrate the efficacy of $P_{P}$ as a useful prior, we measured the Mean Squared Error (MSE) between various features and the high-quality pathology prompt $P_{HP}$ in Figure \ref{fig:pipeline}(a). 
Lower MSE values indicate that the prompt provides semantics closer to those of high-quality images.
We compared $P_{P}$ with prompts from other sources: low-quality pathology prompt $P_{LP}$, extracted by the foundation model from low-quality input images, and $P_{Res}$, extracted from the output images of Restormer.
The results in Figure \ref{fig:restore_prompt_mse} consistently show across different base models that: (1) the low-quality pathology prompt is significantly impacted by degradation, with semantics greatly divergent from $I_{HQ}$, and (2) $P_{P}$ is comparable with semantics after process of Restormer.
These results substantiate the success of our defocus-aware prompt restoration in extracting high-quality semantics from degraded images.

\begin{figure*}[t]
    \centering
    \includegraphics[width=1\linewidth]{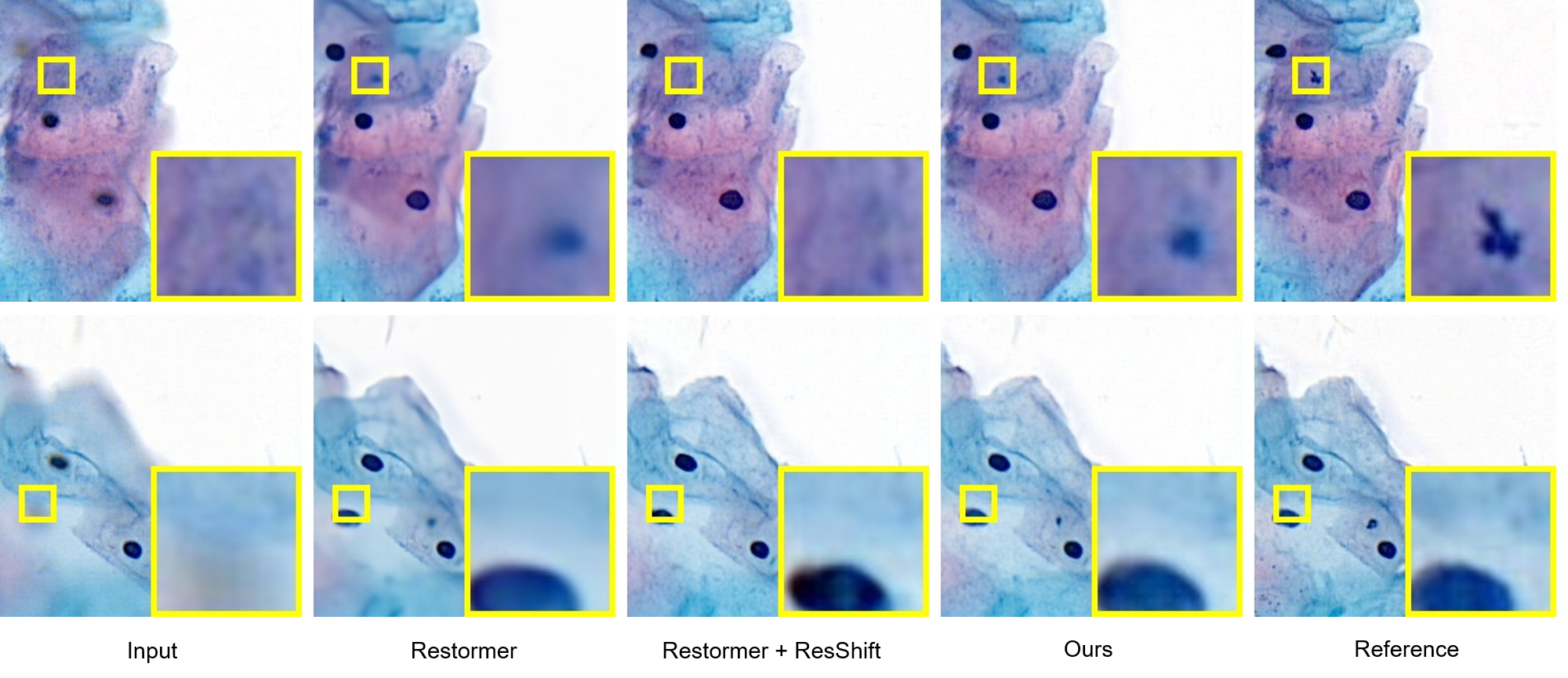}
    \caption{Visual comparison between different methods on 3DHistech. The distinguished regions within the yellow bounding boxes are zoomed in at the bottom right.}
    \label{fig:3d-visualization}
\end{figure*}

\textbf{Visualization on 3DHistech} Fig. \ref{fig:3d-visualization} visualizes the images restored by different methods on 3DHistech. 
Restormer can remove most of the defocus blur but produce over-smooth outcomes. 
ResShift helps produce relatively sharp results but still fails to preserve the cellular structure of the input image.
The proposed method can effectively restore the high-frequency details with excellent structural preservation, which is consistent with the quantitative results.

\begin{table*}[h]
\centering
\resizebox{1.0\linewidth}{!}{
\begin{tabular}{c|ccccccc}
\hline
Method           & Uformer   & GRL    & INIKNet & NRKNet   & RAT              & Restormer         & \textbf{P-former (ours)} \\ \hline
Runtime(s/image) & 0.0359    & 0.1346 & 0.0754  & 0.0182   & 1.2422           & 0.0455            & 0.0943         \\ \hline
Method           & Resfusion & IR-SDE & DA-CLIP & ResShift & Restormer+IR-SDE & \textbf{P-diffusion (ours)} & \textbf{P-former + P-diffusion (ours)}           \\ \hline
Runtime(s/image) & 0.2510    & 7.3398 & 2.2180  & 0.3341   & 7.3853           & 0.4742            & 0.5235    \\     \hline
\end{tabular}
}
\caption{Comparison of inference times across different models.}
\label{tab:time}
\end{table*}

\section{Supplementary Discussion}

\textbf{Runtime.}
Due to variations in tissue area, each WSI contains a different number of patches, making it difficult to provide an exact ‘Runtime Per Slide’. 
The runtime per image of various models is shown in Table \ref{tab:time}. 
We are currently working on integrating our method into existing scanners in the lab.
Our preliminary tests show that P-former for single-layer scanning is already faster than multi-layer scanning, though P-diffusion still has room for optimization.

Beyond speed, we envision leveraging the advantages of single-layer scanning to revolutionize scanner workflows and create a new scanning paradigm: by offloading image processing to the GPU, the scanner can immediately proceed to the next sample, reducing wait times and human intervention. Additionally, eliminating vertical adjustments enables smoother and faster horizontal movement.

\textbf{How to mitigate hallucinations?} Hallucinations during the diffusion process are mitigated through a combination of constraints: the results from P-former provide a solid foundation, the pathology prompt restricts the semantic distribution, and the edge prompt constrains the structural details of the generated image.

\textbf{How to handle the errors in prompts?} 
While prompts of lower quality may not yield optimal results, they still improve performance beyond the baseline by introducing additional information.
In designing our method, we employed attention or gating mechanisms to integrate prompts into the network, effectively filtering useful information and reducing error propagation.
As shown in Table \ref{tab:ablation_stage1}, even unoptimized prompts $P_{LP}$ can enhance performance due to the prior knowledge they incorporate.



\end{document}